\documentclass[conference]{IEEEtran}
\IEEEoverridecommandlockouts
\usepackage{amsmath,amssymb,amsfonts}
\usepackage{algorithmic}
\usepackage{graphicx}
\usepackage{textcomp}
\usepackage{xcolor}
\usepackage{enumerate}

\def\BibTeX{{\rm B\kern-.05em{\sc i\kern-.025em b}\kern-.08em
    T\kern-.1667em\lower.7ex\hbox{E}\kern-.125emX}}

\AtBeginDocument{\colorlet{defaultcolor}{.}} 

\usepackage[citestyle=numeric, backend=bibtex]{biblatex}
\graphicspath{{images/}}
\usepackage{multicol}
\usepackage{soul}  
\usepackage{multirow}
\usepackage{xcolor}

\newcommand{\updated}[1]{{#1}}
\newcommand{\atabey}[1]{{#1}}

\begin{document}

\title{Classification of Stochastic Processes with Topological Data Analysis
\thanks{Computing resources used in this work were provided by the National Center for High Performance Computing of Türkiye (UHeM) under first author's project which grant number 4010242021.}
}

\author{\IEEEauthorblockN{İsmail Güzel}
\IEEEauthorblockA{\textit{Mathematical Engineering} \\
\textit{İstanbul Technical University}\\
İstanbul, Türkiye \\
iguzel@itu.edu.tr}
\and
\IEEEauthorblockN{Atabey Kaygun}
\IEEEauthorblockA{\textit{Mathematical Engineering} \\
\textit{İstanbul Technical University}\\
İstanbul, Türkiye \\
kaygun@itu.edu.tr}
}
\maketitle

\begin{abstract}
    In this study, we examine if engineered topological features can distinguish time series sampled from different stochastic processes with different noise characteristics, in both balanced and unbalanced sampling schemes.  We compare our classification results against the results of the same classification tasks built on statistical \updated{and raw} features.  We conclude that in classification tasks of time series, \updated{different} machine learning models built on engineered topological features perform consistently better than those built on standard statistical \updated{and raw} features.
\end{abstract}


\section{Introduction}

    Topological Data Analysis (TDA) is a new field of data science that uses topological and geometric tools to infer relevant features from potentially complex data. TDA has developed procedures that are not based
    on traditional statistical or machine learning algorithms, and its methods are now used in a variety of fields from finance~\cite{majumdar2020clustering} to medicine~\cite{nanda2014simplicial}. 
    
    We concentrate on engineered topological features using persistent homology. The information that persistent homology yields on the change of topological features of a given point cloud can be presented in various different ways such as barcodes~\cite{ghrist2008barcodes}, persistence diagrams~\cite{cohen2007stability}, landscapes~\cite{bubenik2015statistical},
    images~\cite{image2017persistence}, terraces~\cite{terrace2018persistence},
    entropy~\cite{merelli2015topological} and curves~\cite{chung2019persistence}. Persistent homology has become increasingly important for the analysis of noisy signals and time series in a variety of domains. It has been used to solve problems in signal processing and systems engineering, to quantify periodicity in the literature~\cite{perea2015sliding}, for clustering tasks~\cite{seversky2016time}, for classifying tasks~\cite{umeda2017time}, or to detect early signals for critical transitions~\cite{gidea2017topological, gidea2018topological}.   

    The most commonly encountered problem in TDA applications is that persistent homology is computationally expensive. Most popular libraries used for TDA calculations often resort to parallel computing~\cite{,burella2021giottoph} and/or off-loading heavy computations to GPUs~\cite{zhang2020gpu} to alleviate the problem.

    In this study, we investigate whether one can observe any topological differences between time series samples from different stochastic processes such as Wiener and Cauchy. First, in Section~\ref{sect:background} we introduce our framework we based on persistent homology. Next, we introduce the topological and statistical features we are going to use in Section~\ref{sect:featureengineering}. Finally, in Section~\ref{sect:experiments} we apply various machine learning classification algorithms on the features we engineered.

\section{Background}\label{sect:background}
    
    In this Section, we briefly introduce the necessary background for the later Sections.
    
\subsection{Stochastic Processes}
    Stochastic processes have been used to model systems that evolve probabilistically through time in a variety of fields such as  finance~\cite{kijima2002stochastic}, physics~\cite{jacobs2010stochastic}, and image-signal processing~\cite{won2004stochastic}, etc. 
    
    Levy process family is a vast family of stochastic processes with applications in insurance risk modelling~\cite{kluppelberg2004ruin, landriault2021analysis}, finance~\cite{schoutens2003levy,tour2018cos}, and economics~\cite{geman2002pure}. A Levy process is a continuous-time stochastic process that has stationary independent increments with three main components: the deterministic component, the Brownian motion component, and a measure quantifying the rate at which discrete jumps occur. 
    
    Here, we give two process from the family of Levy processes as follows:

\subsubsection{Wiener Process}

	A standard Wiener process (often called Brownian motion) is a random variable $X_t$ that depends continuously on $t \in [0,T]$ and satisfies the following:
	\begin{itemize}
		\item For $0 \le s < t \le T$,
		\begin{displaymath}
			X_t - X_s \sim \sqrt{t - s} N(0,1),
		\end{displaymath}
		where $N(0,1)$ is the normal distribution with zero mean and unit variance with $X_0 = 0$.  
		\item For $0 \le s < t < u < v \le T$, $X_t-X_s$ and $X_v-X_u$ are independent.
	\end{itemize}

\subsubsection{Cauchy Process}

	The Cauchy process $X=(X_t,t \geq 0)$ is a process with state space $\mathbb{R}$ and the stationary independent increments such that 
	$ X_{t+s} - X_s $ has the Cauchy probability density function 
	\begin{displaymath}
		\rho_t(x) = \frac{1}{\pi}\frac{t}{t^2 + x^2}.
	\end{displaymath}
    
    Actually, the Wiener process is a simple case of Levy process. It has the independent and identically distributed Gaussian increments with variance equal to increment length. The Cauchy process, on the other hand, is a Brownian motion with a Levy subordinator using the location parameter 0 and the scale parameter $ \frac{t^2}{2} $, over the maximum time interval $ t $.\atabey{ We refer readers to Sec. 1.3 of~\cite{applebaum2009levy} for details about the Levy process and its subordinators.} So, the difference between the Brownian motion and Cauchy process is that taken the Levy subordinator. We will try to examine this difference whether any statistically or topologically differences explain.

\subsection{Takens' Delay Embedding for Time Series}\label{TakensEmbed}
	
	Time series do not naturally have point cloud representations. We are going to use Takens' Embedding Theorem~\cite{takens1981detecting} to convert a time series to a point cloud before we can begin applying TDA methods. Given a times series $ x_t $ for $ t= 1,2,\ldots,T $, we convert this time series into a point cloud with points $ v_i = (x_i, x_{i+\tau}, \ldots, x_{i+(d-1)\tau})^{T}$, where $ \tau $ denotes a delay parameter, and $ d $ is the dimension of the space the point cloud embedded into. 

	Takens's embedding depends on two parameters $d$ and $\tau$, and must be determined in practice. Pereira and de Mello~\cite{pereira2015persistent} used the first minimum of the mutual information between the signal and its time delayed version to determine the time deal $ \tau $. Many studies ~\cite{truong2017exploration, khasawneh2016chatter} used the false nearest neighbor method~\cite{FalseNearestNeighbor} to determine the embedding dimension $ d $. So, we will use the same methods to determine the parameters $\tau$ and $d$ on our experiments. 
    
\begin{figure}[ht]
		\centering
		\includegraphics[scale=0.275]{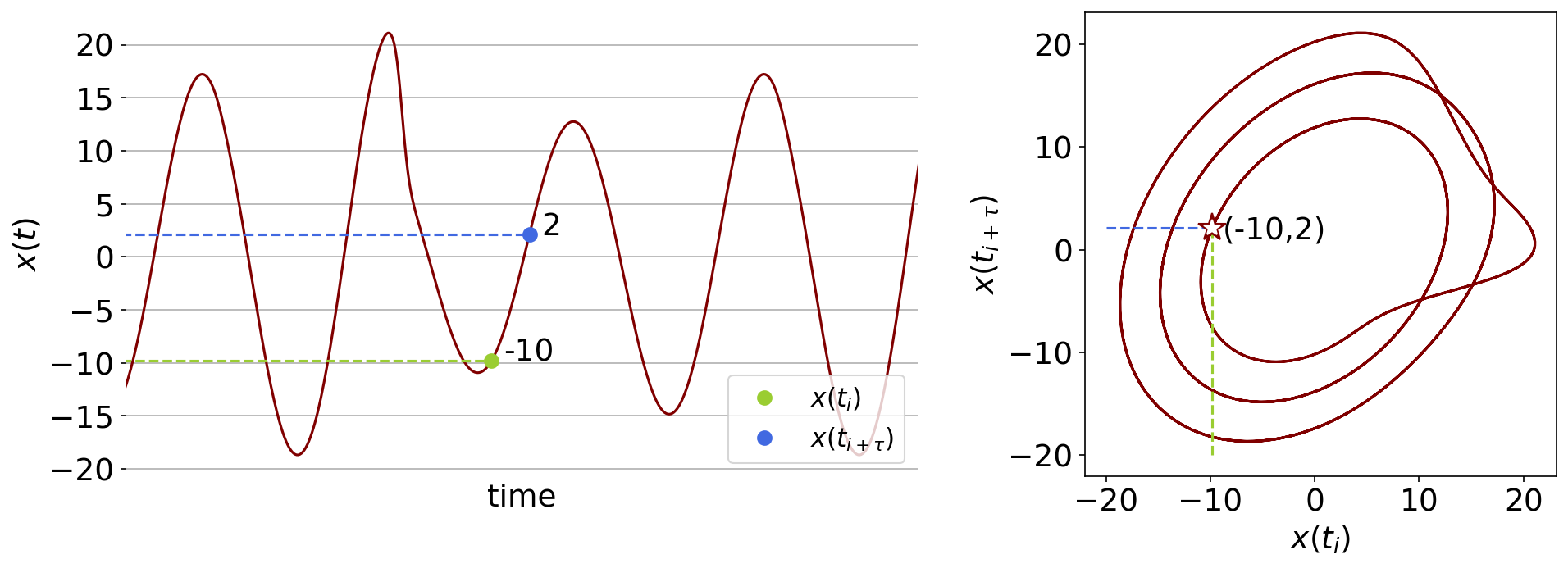}
		\caption{The delay embedding of time series as $x$ variable of R\"{o}ssler system. The delay parameter $\tau = 25$ is choosen by utilize the mutual information while the embedding dimension $d$ is selected 2 because of visualization. }\label{timedelay}
	\end{figure}

    
    The delay embedding guarantees the preservation of topological features of a time series but not its geometrical properties.
    Even if it causes the loss of some geometric properties, it still has an important role. For instance, in industrial sensor data applications, sometimes one has just one variable to synthesize the whole system. For this purpose, the delay embedding allows one to reconstruct attractors of an unknown system. For example, the attractor of the Rössler system~\cite{venkataraman2016persistent}, 
    is shown in Figure~\ref{timedelay}.
    
\subsection{Topological Data Analysis}\label{sect:TDA}
	
	One of the main goals of algebraic topology is to classify topological spaces based on their features. In data science, on the other hand, extracting important features from large datasets to classify or cluster these extracted features appears as a recurrent problem. TDA aims to bridge topology and data science from this perspective. 
	
\subsubsection{Persistent Homology}
	
	One important tool that TDA employs in the process is called persistent homology. Persistent homology of a point cloud gives us engineered features depending on a fixed natural number $ n $. For instance, the results obtained for $ n=0,1 $ and $ 2 $ indicate the number of connected components, loops and 2-dimensional voids in a given data set.
    
    To get persistent homology from point clouds, one needs a filtered simplicial complex such as the Alpha complex, the $\check{C}$ech complex, or the Vietoris-Rips complex. We refer interested readers to \cite{carlsson2009topology} for an introduction to persistent homology, and more mathematical details on building topological spaces from point clouds.

\subsubsection{Barcodes}

\begin{figure}[ht]\label{figure:1}

	\centering
	\includegraphics[scale=0.55]{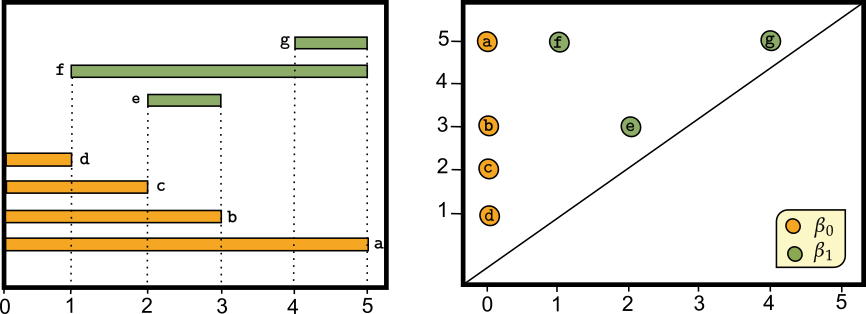}
	\caption{The barcodes and persistence diagram.}
    \end{figure}

    Persistent homology of a point cloud is typically represented by a \emph{barcodes} or a \emph{persistence diagram}. (See Figure~\ref{figure:1}.) 

	Each bar in the barcodes corresponds to a homology class of the parameterized family of topological spaces obtained from the point cloud. The left end point of each bar corresponds to the parameter value at which the topological feature first appears, also referred to as the \emph{birth time} $b_i$, and the right end point corresponds to the parameter value at which the topological feature disappears, referred to as the \emph{death time} $d_i$. Thus, the bars in a barcode diagram can be represented as a collection of pairs $ (b_i,d_i) $. 
	
\subsubsection{Persistence Diagrams}
	
    Given a barcode $\{(b_i,d_i)\mid i\in I\}$ as pairs of birth and death times, when we plot these pairs $ (b_i,d_i) $ in $ \mathbb{R}^2 $ we get the corresponding \emph{persistence diagram}. Note, $ b_i \leq d_i $ for all $ i $ so all points in a persistence diagram lie on or above the diagonal $ y = x $. 
    
\subsubsection{The Wasserstein and The Bottleneck Distances}

\begin{figure}[ht]
		\centering
		\includegraphics[scale=0.45]{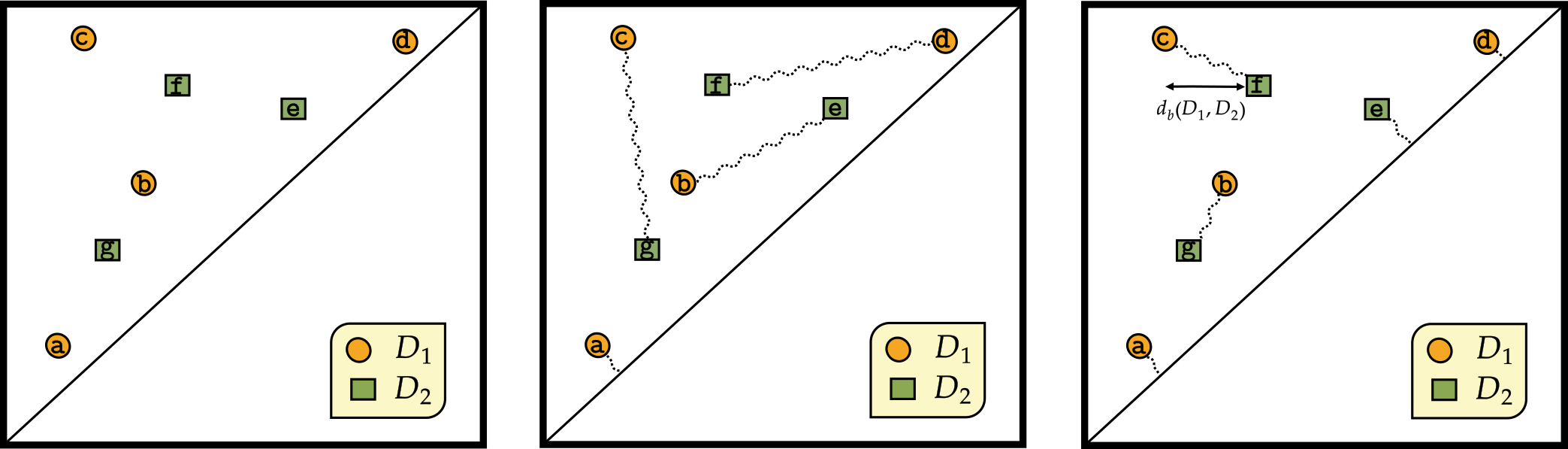}
		\caption{The best matching on given two persistent diagram and the bottleneck distance of given two diagrams $D_1$ and $D_2$. }\label{bestmatching}
	\end{figure}

    Given two persistence diagrams $ D_1$ and $D_2$, and , we have bijections (perfect matching) of the form $ \varphi: D_1 \rightarrow D_2 $ as seen in Figure~\ref{bestmatching}. The \emph{Wasserstein distance}~\cite{kerber2017geometry} of $D_1$ and $D_2$ is defined to be 
    \[ W_{p}(D_1, D_2)=\inf _{\varphi}\left(\sum_{x \in D_1}\|x-\varphi(x)\|_{\infty}^{p}\right)^{1 / p}. \]
    We also satisfy the points on the diagonals $ \Delta = \{(s,s) | s\in \mathbb{R}\} $ of $D_1$ and $D_2$ to ensure a perfect matching always exists. Because of the diagonal $ \Delta$, the diagrams do not have to have same the number of points to calculate the Wasserstein distance.
	
	The Wasserstein distance is called the \emph{bottleneck distance} if we let $ p \rightarrow \infty $,
	\[ d_B(D_1, D_2) = \inf_{\phi} \sup_{x\in D_1}||x-\phi(x)||_{\infty}. \]

\subsubsection{Persistence Landscapes}
	
	Given a persistence diagram $\{(b_i,d_i)\mid i\in I\}$, after \cite{bubenik2020persistence}, one can construct the corresponding persistence landscape as a sequence of piece-wise linear functions, $\lambda_{1}, \lambda_{2}, \ldots: \mathbb{R} \rightarrow \mathbb{R}$ with slopes $0$, $1$, or -1.  
    
	Here we let
	$$
	\lambda_{k}(t)=\mathrm{kmax}\left\{f_{\left(b_{i} d_{i}\right)}(t)\right\}_{i \in I},
	$$
	where kmax is the $ k $-th largest element and 
	$$
	f_{(a, b)}(t)=\max (0, \min (a+t, b-t)).
	$$
	
\begin{figure}[ht]
		\centering
		\includegraphics[scale=0.35]{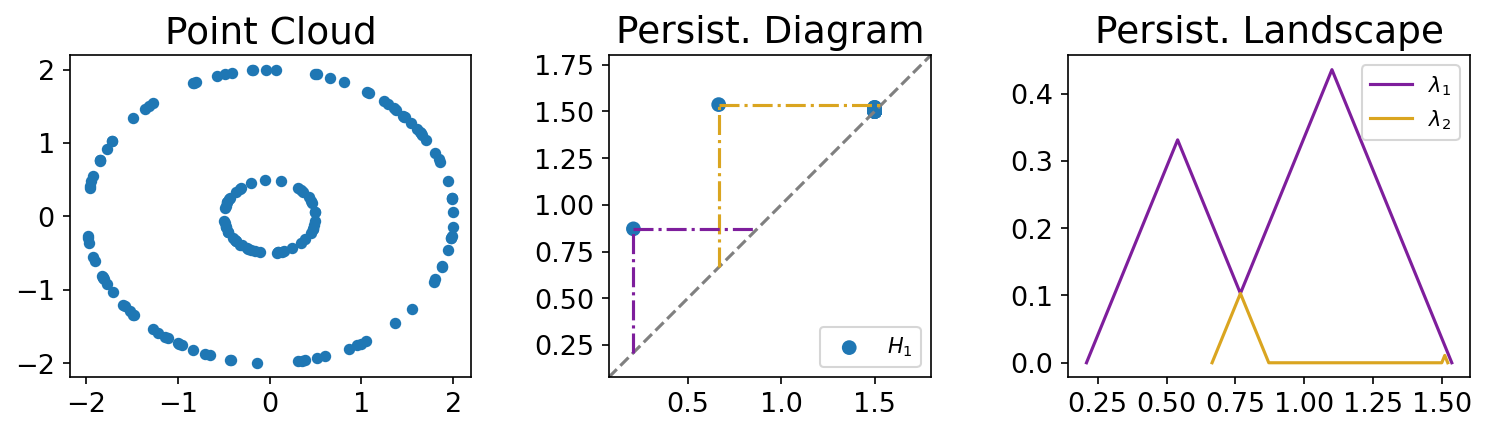}
		\caption{The persistence diagram and corresponding first two persistence landscapes for 1 dimensional homology classes for the intertwined two circles. }\label{landscapes}
	\end{figure}

	For example, we suppose the point cloud sampled from the two intertwined circles shape in Figure~\ref{landscapes}. To investigate the topological features of that point cloud, one can consider the number of local maxima of the persistence landscapes or the area of under that curve. As seen in Figure~\ref{landscapes}, there are two local maxima of the first persistence landscapes of a given point cloud.

\subsubsection{Betti Curves}
    Let $ D=\{(b_i,d_i)\mid i\in I\} $ be a persistence diagram. The Betti curve of diagram $D$ is the function $ \beta_D : \mathbb{R} \rightarrow \mathbb{N} $ whose value on $ s\in\mathbb{R} $ is the number of points $ (b_i,d_i) $ in $ D $ such that $ b_i\leq s < d_i $. More formally, it can be given as 

    \[ \beta_D(s) : \#\{(b_i,d_i) \in D \mid b_i \leq s < d_i \}. \]
	
\begin{figure}[ht]
	    \centering
	    \includegraphics[scale=0.5]{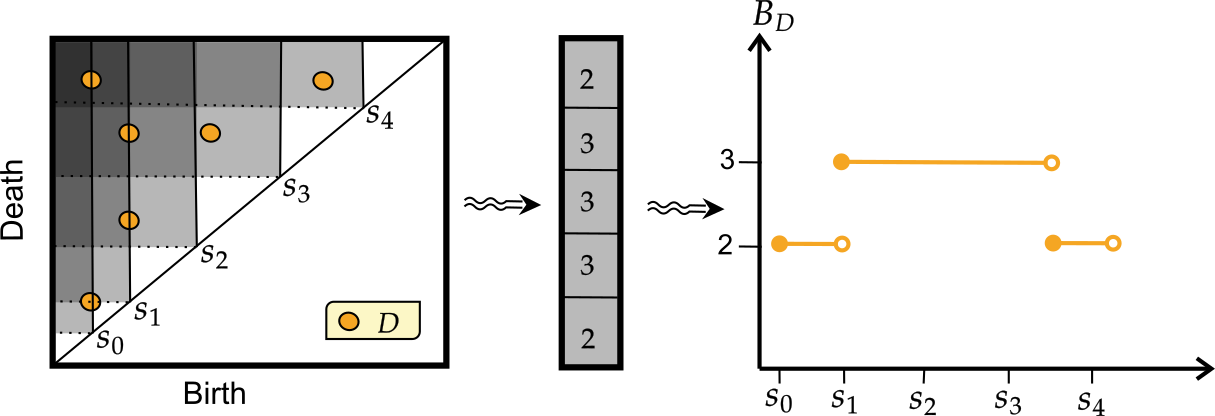}
	    \caption{From the persistence diagram to the corresponding Betti curve. For example, there are three orange pairs in the shaded region on the persistent diagram at $s_3$ while two orange pairs are at $s_4$.  }\label{betticurve}
    \end{figure}

	An example of the calculating Betti curves from a given persistence diagram is shown in Figure~\ref{betticurve}. Shaded gray region is the persistence diagram region containing pairs to contributing $\beta_D(s)$.

\section{Feature Engineering}\label{sect:featureengineering}

    Feature engineering plays an important role on classification and clustering tasks in many machine learning models. In this section, we outline the details of the engineered featured our analyses are based on. 

\subsection{Statistical Features}{\label{subsec:statistical}}
    Many time series classification tasks require a number of preprocessing steps. One of the most straightforward preprocessing steps on time series is to create descriptive statistical features such as the mean, the variance, and the entropy. For this study, we chose the following list of descriptive statistical features as follows:
    \begin{multicols}{3}
    	\begin{enumerate}
    		\item Mean
    		\item Variance
    		\item Entropy
    		\item Lumpiness
    		\item Stability
    		\item Hurst
    		\item Std 1st-der
    		\item Linearity
    		\item Binarize mean
    		\item Unitroot KPSS
    		\item Histogram mode
    	\end{enumerate}
    \end{multicols}

\subsection{Topological Features}{\label{subsec:topological}}

    There are three popular indirect methods to use persistence diagrams with machine learning algorithms. Firstly, one may consider kernel methods in conjunction with the distance metrics on persistence diagrams. Secondly, one may extract some descriptive statistical features from persistent diagram such as Adcock-Carlsson coordinates, the lifetime, the persistence entropy, etc. Lastly, one may transform persistence diagrams space to a Hilbert space using persistence landscapes, or Betti curves. We refer the reader \cite{otter2017roadmap, chazal2021introduction} for more details about the road map on the topological tools to use on machine learning task.
    
    Let $ D =\{(b_i,d_i)\mid i\in I\} $ be the persistence diagram of a point cloud coming from Taken's Embedding of a time series with the optimal parameters time delay $\tau$ and dimension $d$ as in Section~\ref{TakensEmbed}. Then the topological features can obtained as follows:

\subsubsection{Wasserstein and bottleneck distances}

    We can obtain two new topological features by using Wasserstein and bottleneck distances between given diagram $ D $ and the diagonal diagram $ \Delta $. In the case of the diagonal diagram we get  
    \[ W_1(D,\Delta) = \sum_{i\in I} \frac{d_i-b_i}{2} \mbox{ and } d_B(D,\Delta) = \sup_{i\in I}(\frac{d_i-b_i}{2}).\]  

\subsubsection{Adcock-Carlsson Coordinates}

    Let $ d_{max} $ be the maximum death time over all pairs, and let $ \ell_i = d_i-b_i $ be the life time of topological features. In \cite{adcock2016ring}, the authors used the following summaries of persistence diagrams. 
    	\begin{enumerate}[(a)]
    		\item $ f_1(D) = \sum_{i\in I} b_i\ell_i $
    		\item $ f_2(D) = \sum_{i\in I} (d_{max}-d_i)\ell_i $
    		\item $ f_3(D) = \sum_{i\in I} b_i^2\ell_i^4 $
    		\item $ f_4(D) = \sum_{i\in I} (d_{max}-d_i)^2\ell_i^4$
    	\end{enumerate}
    Using these features we obtain four new features from persistence diagrams of persistent homology in degree 1.

\subsubsection{Persistence Entropy}
	The persistence entropy of the persistence diagram $D$ is defined by
	\[ E(D) = -\sum_{i} \frac{\ell_i}{L_D}\log\left(\frac{\ell_i}{L_D}\right), \]
	where $ L_D = \sum_{i\in I} \ell_i $.
	This adds one more feature for our analyses.

\subsubsection{$L_1$ Norms}

    Let $\beta_D$ be the corresponding Betti curve, and let $\lambda_1$ be the corresponding persistence landscpae. Since the Betti curve and the persistence landscape are an integer-valued and a real-valued functions, respectively, we can calculate their $ L_p $ norms. In practice, we simply use $ p=1 $ for the features of a given persistence diagram and on top of the statistical features we will have two new features using the norms $\|\beta_D\|_1$ and $\|\lambda_{1}\|_1$ 
    
    Therefore, we have summarized with 9 topological features and 11 statistical features to use in classification methods for a time series.
\section{Experimental Contributions}\label{sect:experiments}

    All implementations are done via three libraries of the python programming language~\cite{python3}. For sampling the stochastic processes, we used the {\tt stochastic} package~\cite{stochastic}. We obtained the topological features and persistence diagrams from point cloud using the {\tt giotto} and the {\tt giotto-ph} libraries~\cite{tauzin2020giotto, burella2021giottoph}. We extracted the statistical features from time series using the \textit{Kats} library~\cite{kats}.

\subsection{Results with Balanced Sampling}

\subsubsection{Simulations}

\begin{figure}[ht]\label{fig:topological}
	\centering
	\includegraphics[scale=0.23]{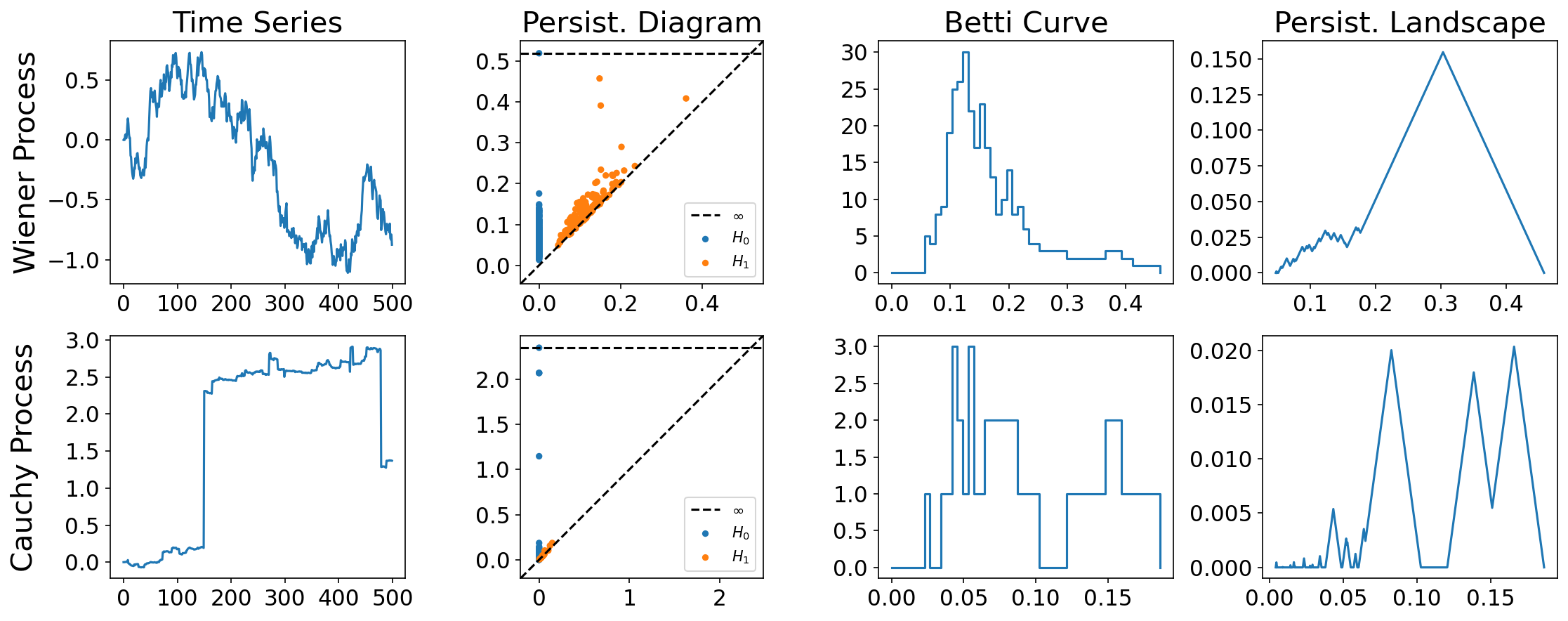}
	\caption{An example of the persistence diagram, Betti curve and persistence landscapes which obtained from Wiener and Cauchy process.}
	\end{figure}

	We generated 1000 time series with the same length of 500 time steps for both Cauchy and Wiener processes on the time interval $ [0,2] $. So, in total we have 2000 time series and their corresponding labels as `Cauchy' and `Wiener.' We then generated statistical features for each time series as described in Section \ref{subsec:statistical}. Next, we obtained the topological features for each of the time series as described in Section \ref{subsec:topological}. For the topological features, we focus on only degree 1 persistence diagram since these can capture any periodicity behavior that may appear in the time series. For the persistence diagram corresponding to each time series, we obtain persistence entropy, bottleneck distance, Wasserstein distance, the four Adcock-Carlsson coordinates, and the $ L_1 $ norms of Betti curve and first persistence landscape. (See Figure~\ref{fig:topological}.)

\subsubsection{Correlations and distances between features}	

\begin{figure}[ht]
	\centering
	\includegraphics[scale=0.35]{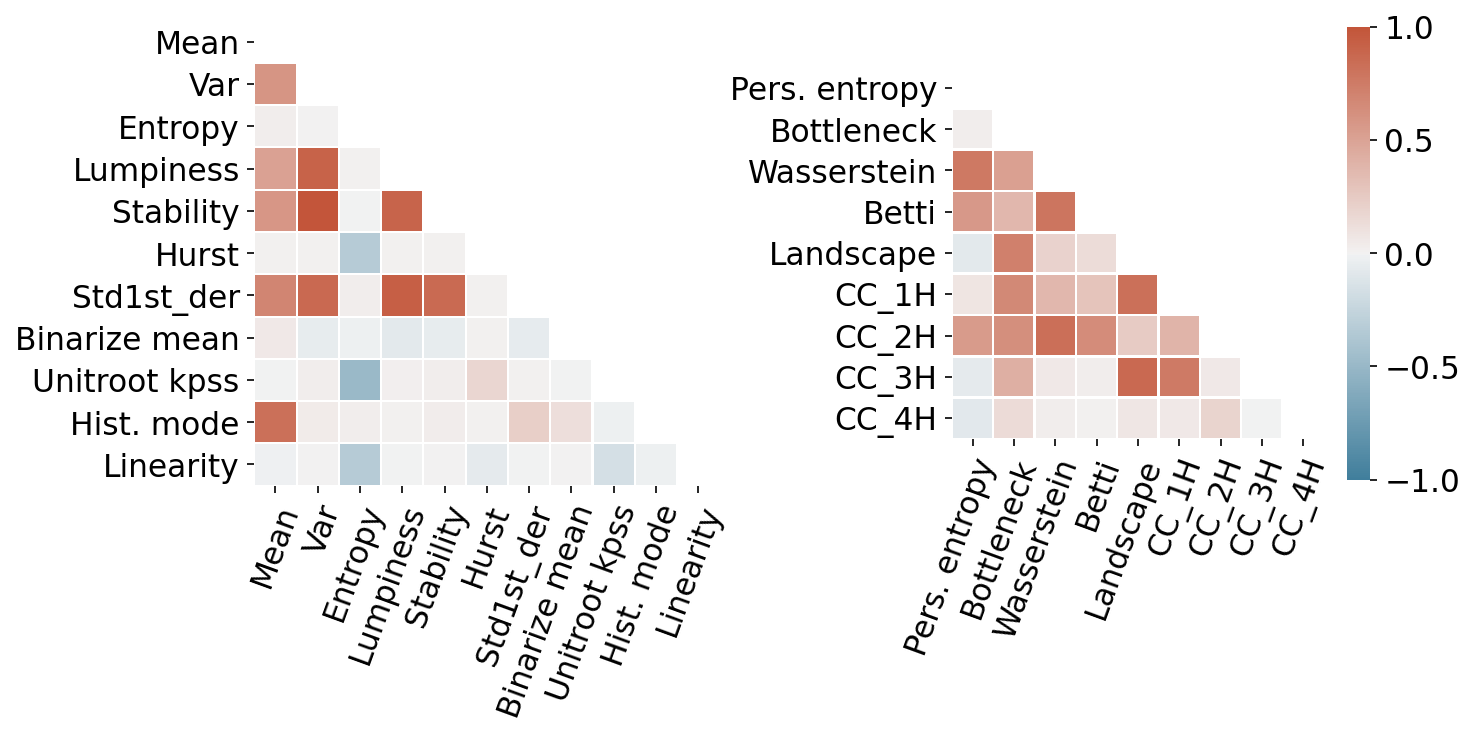}
	\caption{The correlation within themselves for the statistical (left) and topological (right) features.}
	\label{correlation}
	\end{figure}

	For both the statistical and the topological features, we display the correlation coefficients in Figure~\ref{correlation} as a heatmap. Whereas the statistical features appear uncorrelated, the topological features are mostly correlated since they all come from persistence diagrams.
	
\begin{figure}[ht]
	\centering
	\includegraphics[scale=0.35]{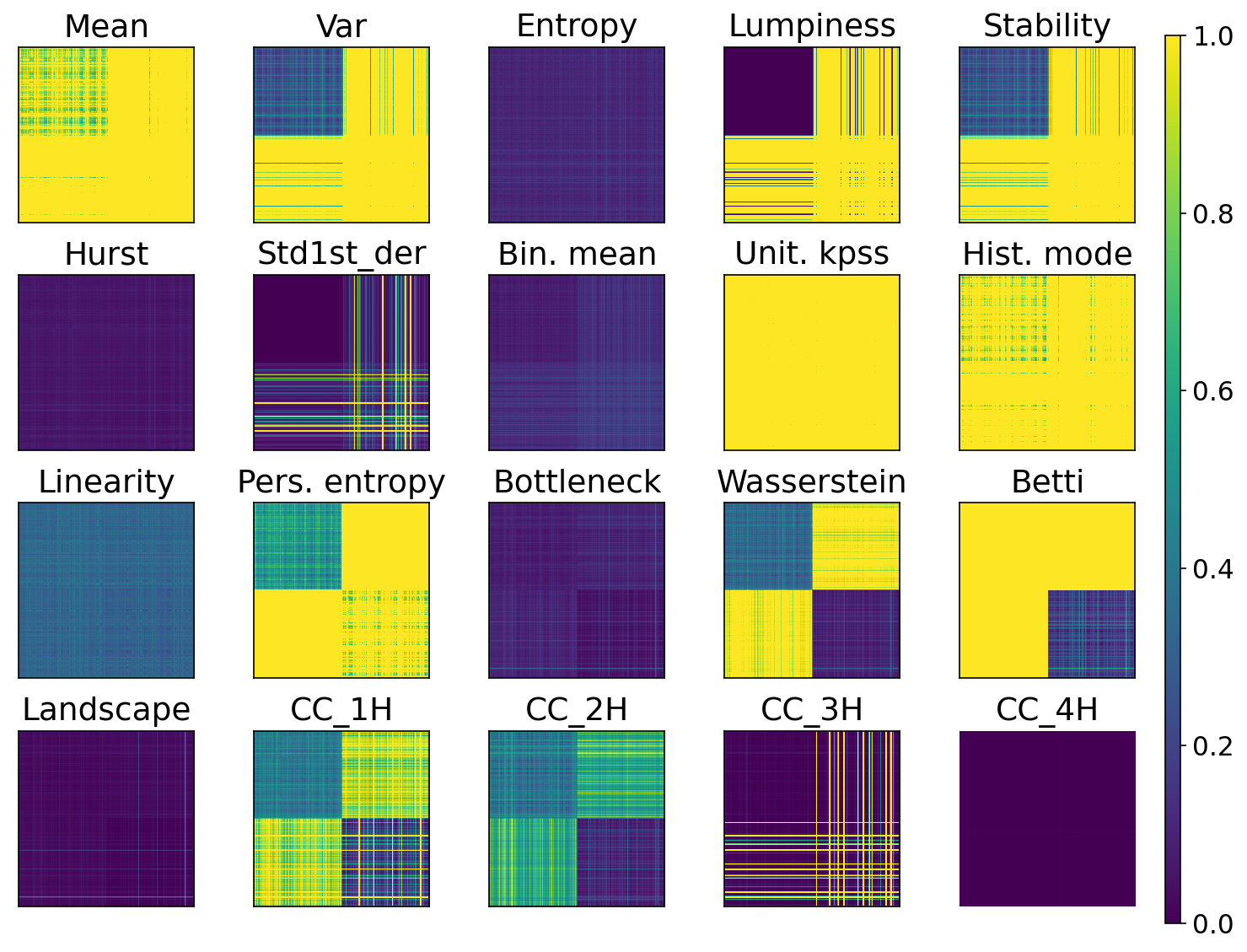}
	\caption{The pairwise distances between the two classes of time series using topological and statistical features. Balanced sampling results.}
	\label{pairwise}
	\end{figure}

    To distinguish the features, the pairwise distances of the two classes of time series in all features are given in Figure~\ref{pairwise}, also as a heatmap. 
	
	One can see that the Wasserstein distance, the Adcock-Carlsson coordinates 1 and 2 can distinguish between each group in Figure~\ref{pairwise}. The distance matrices indicate that the individual clusters show small intra-cluster distances with larger inter-cluster distances on the topological features.

\subsubsection{Classification}

	After simulation and featurization,\atabey{ we can now apply machine learning models to our datasets that consists of raw, statistical and topological features with labels. We used Logistic Regression (LGR), Decision Tree (DCT), k-Nearest Neighbor (KNN), Random Forest (RFT), Support Vector Classifier (SVC), Multi Layer Perceptron (MLP), Linear Discriminant Analysis (LDA) and XGBoost (XGB). We used python library sklearn~\cite{sklearn} and xgboost~\cite{chen2016xgboost} with default parameters.} Before we start building models, we split our synthetic dataset into the train and the test datasets with sizes 80\% and 20\%, respectively.  We also did $ 5 $-fold cross validations for each model over the whole dataset.  

	\begin{figure}[ht]
		\centering
		\includegraphics[scale=0.3]{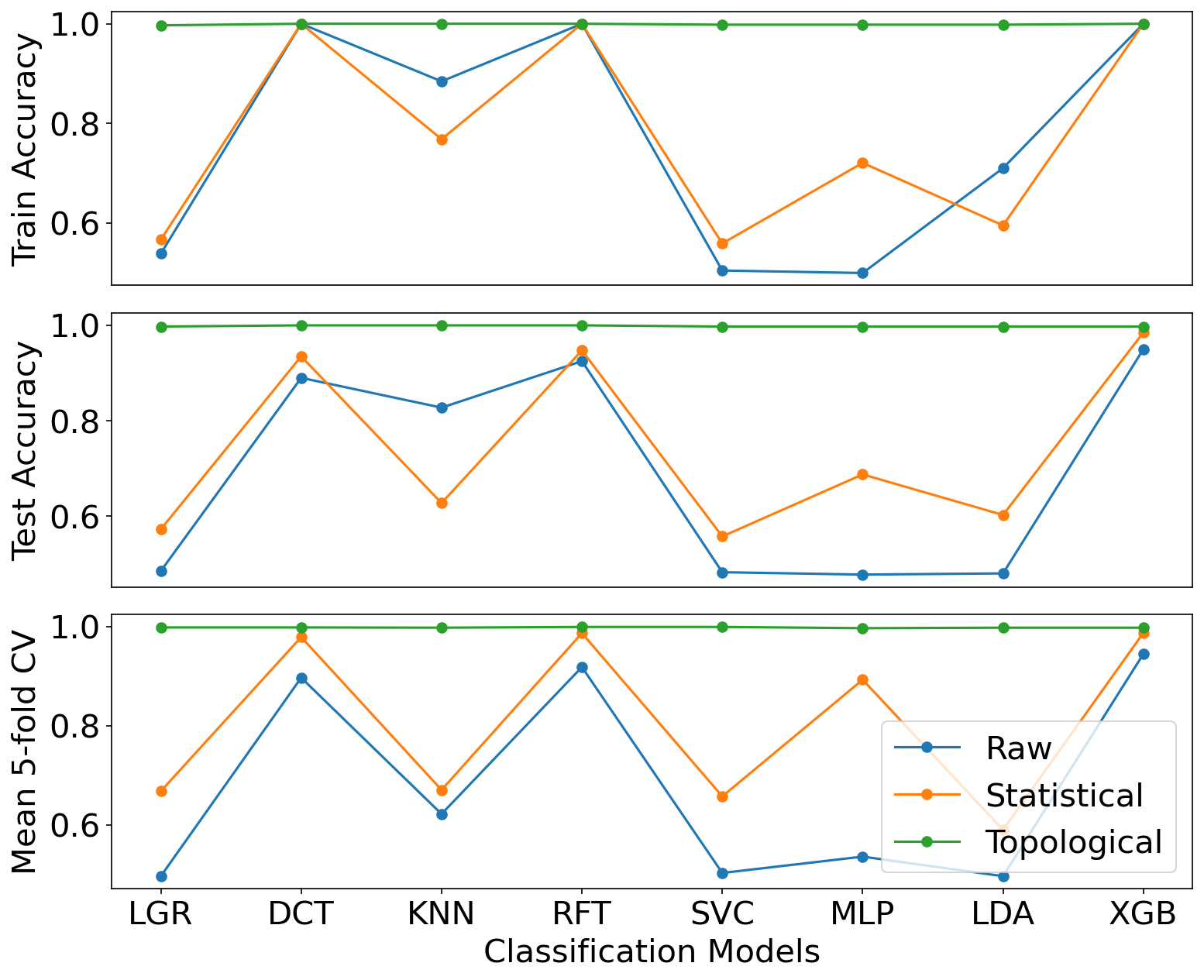}
		\caption{The cross validation results for machine learning classification models based on topological, statistical features, and \updated{raw features} for the balanced dataset. }
		\label{MLModels}
	\end{figure}

	\begin{figure}[ht]
		\centering
		\includegraphics[scale=0.32]{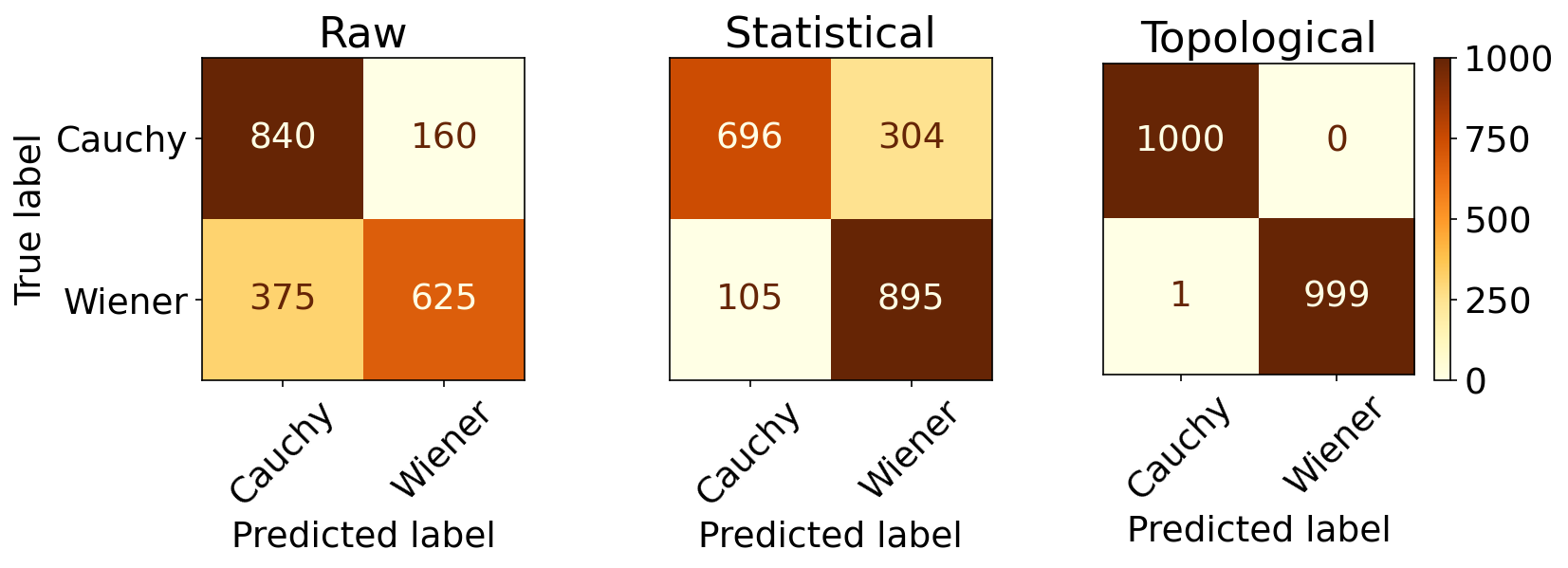}
		\caption{The confusion matrices from the \updated{raw features}, the statistical features and the topological features for the unbalanced dataset from the KNN algorithm.}
		\label{confusionmatrix}
	\end{figure}
	
	As one can see in Figure~\ref{MLModels}, the result shows that models built on topological features consistently show better accuracies than the models built on statistical \atabey{and raw features. The results of 5-fold cross-validation indicate that the most performant representative features, in order, are topological, statistical and raw features, respectively. One can conclude from these results that the feature engineering plays a critical role since the results on engineered features are better than the results on the raw datasets. }

	\atabey{When we restrict to statistical and raw features the SVC, LDA and LGR models are relatively less performant compared to other models while the XGB model is the most performant on all features. On the other hand, it appears that DCT model over-fit on the raw and statistical features for the train set. The confusion matrix for the k-nearest neighbor (KNN) model for the balanced dataset, which is the worst performing model for all features, is shown in Figure~\ref{confusionmatrix}.}
	
\begin{figure}[ht]
		\centering
		\includegraphics[scale=0.23]{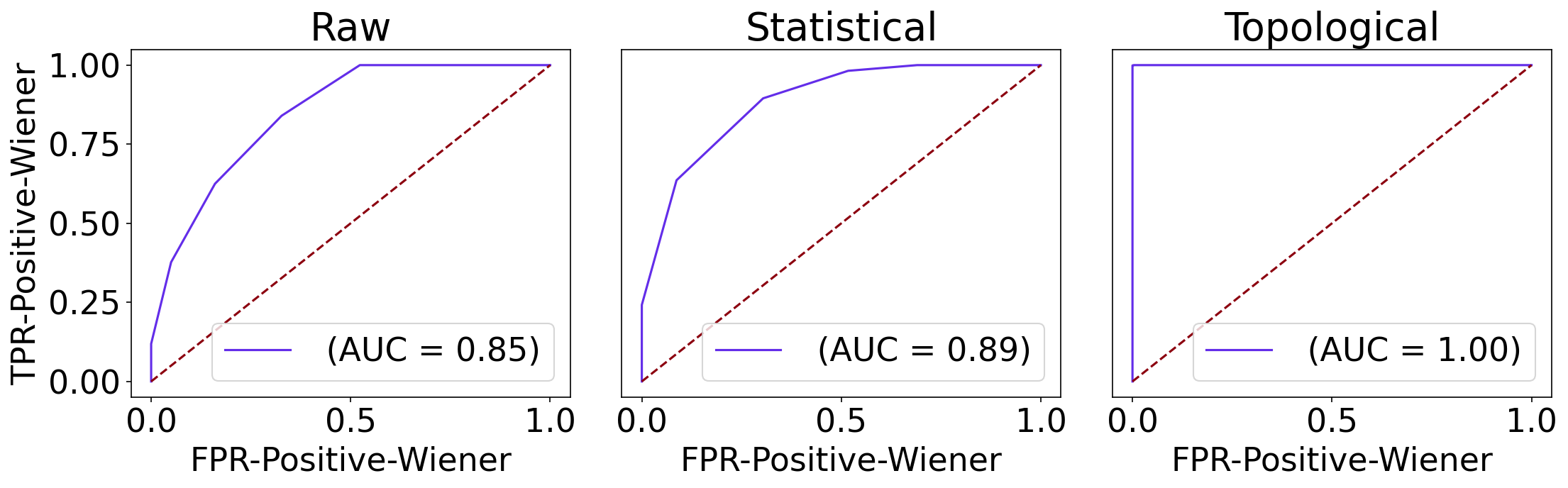}
		\caption{The ROC curve from the statistical features (left) and the topological features (right) for the balanced dataset, from the KNN algorithm.}\label{roccurve}
	\end{figure}

	We use the Receiver Operating Characteristic (ROC) curve to analyze classifier performances. The ROC curve depicts the trade-off between the true positive rate and the false positive rate, or sensitivity vs specificity, for different thresholds of the classifier output. Figures~\ref{roccurve} show the ROC curve for different features with the KNN algorithm. The area under the ROC curve (AUC) indicate the models based on topological features alone show better performance that the models based on the statistical \updated{and the raw} features alone.
	
	\subsection{Results with Unbalanced Sampling}
	
\begin{figure}[ht]
		\centering
		\includegraphics[scale=0.33]{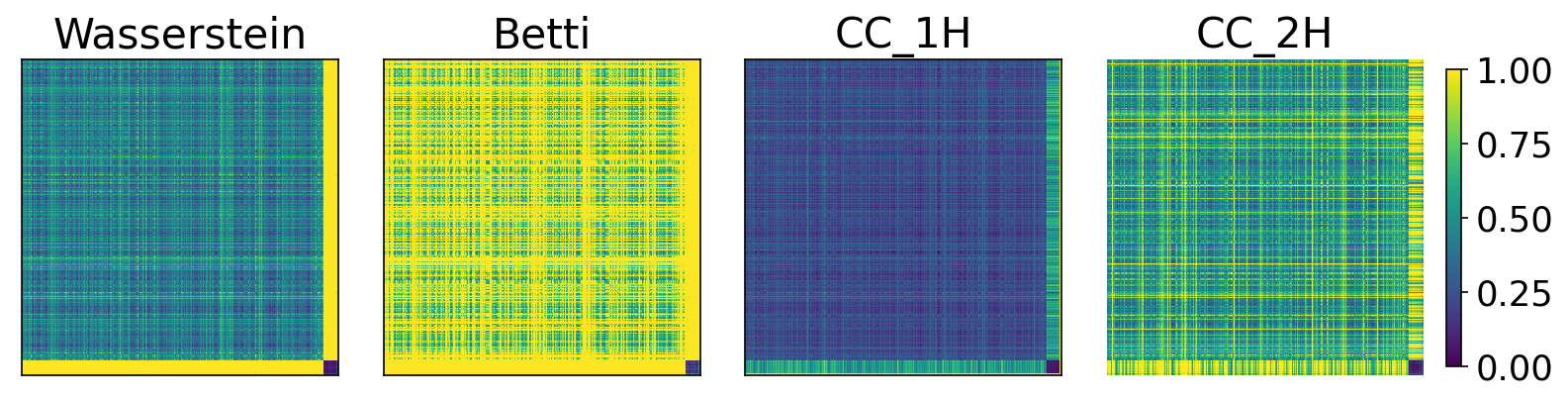}
		\caption{The pairwise distances between the two classes of time series using topological features. Unbalanced sampling results.}
		\label{SelectedFeatures}
	\end{figure}

\begin{figure}[ht]
		\centering
		\includegraphics[scale=0.3]{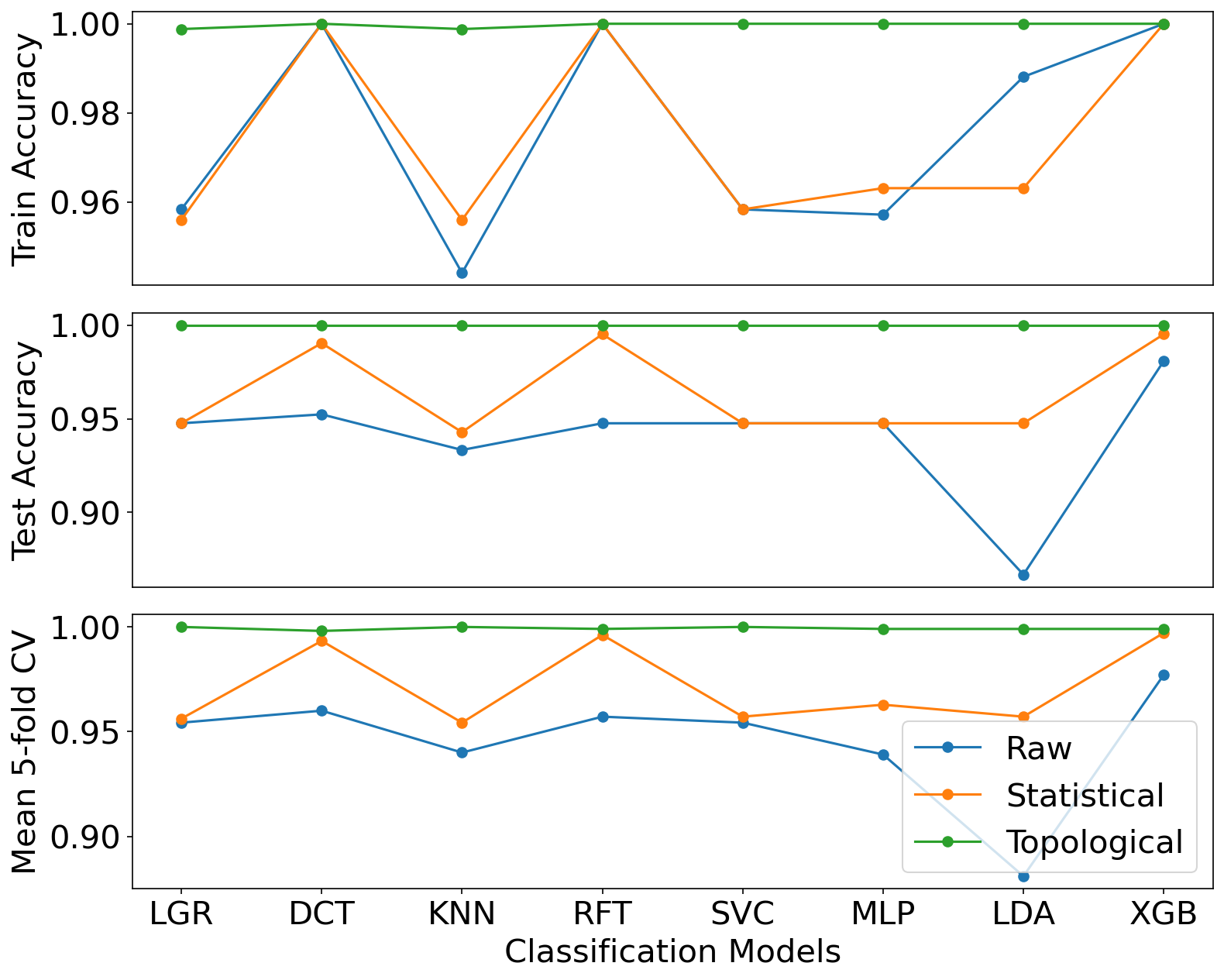}
		\caption{The cross validation results for various classification machine learning models base on topological, statistical features for the unbalanced dataset.}
		\label{MLModels_un}
	\end{figure}

	Most supervised machine learning models require a balanced dataset for the training phase of model building, and perform rather poorly if the dataset is unbalanced in favor of one class. In order to test the performance of our topologically engineered features, we repeated the same processes with an unbalanced sampling of the stochastic processes we are interested in. To show that our proposed approach performs well with unbalanced datasets, \atabey{we generated 50 samples from a Cauchy process and 1000 observations from a Wiener process.} To weigh the importance of the topological features, we look at the distance matrix between engineered topological features by following the same procedure as in balanced dataset case displayed in Figure~\ref{SelectedFeatures}. \atabey{We again observe that the Wasserstein, Betti and Adcock-Carlsson coordinates again distinguish two classes from each other.} After applying the same validation steps (single train/test and 5-fold cross validation schemes), we still observe that the topological features consistently show better accuracies on unbalanced datasets. The results are shown in Figure~\ref{MLModels_un}.
	
\begin{figure}[ht]
		\centering
		\includegraphics[scale=0.3]{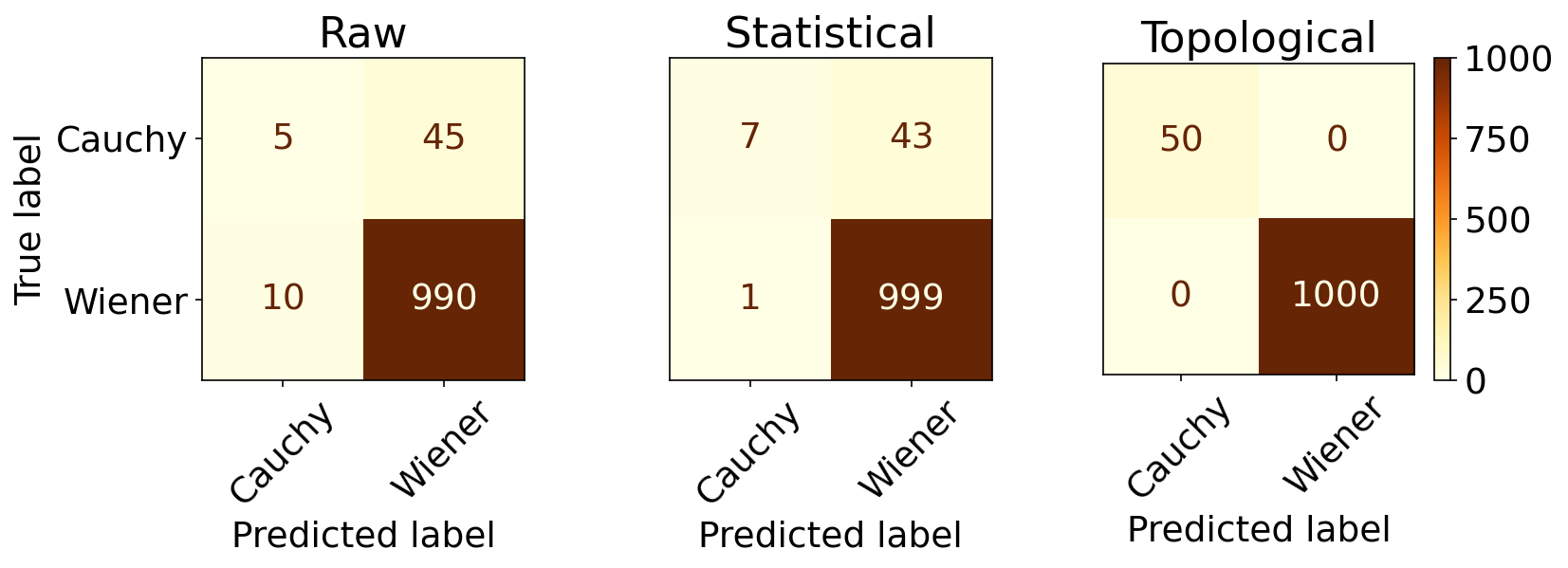}
		\caption{The confusion matrices from the \updated{raw features}, the statistical features and the topological features for the unbalanced dataset,  from the KNN algorithm.}
		\label{confusionmatrix_unbalanced}
	\end{figure}

	For the unbalanced dataset, the confusion matrix of the KNN algorithm on both features is shown in Figure~\ref{confusionmatrix_unbalanced}. We chose the KNN algorithm because for balanced and unbalanced sampling experiments it gave us the worst accuracy results.
		
    \atabey{
    In the unbalanced scheme, we sampled the stochastic processes with varying distributions of classes where the minority class varied between 1\% to 20\% of the majority class. In all experiments, the majority class has a sample size of 1000 observations. 
    }

\begin{figure}[h]
		\centering
		\includegraphics[scale=0.35]{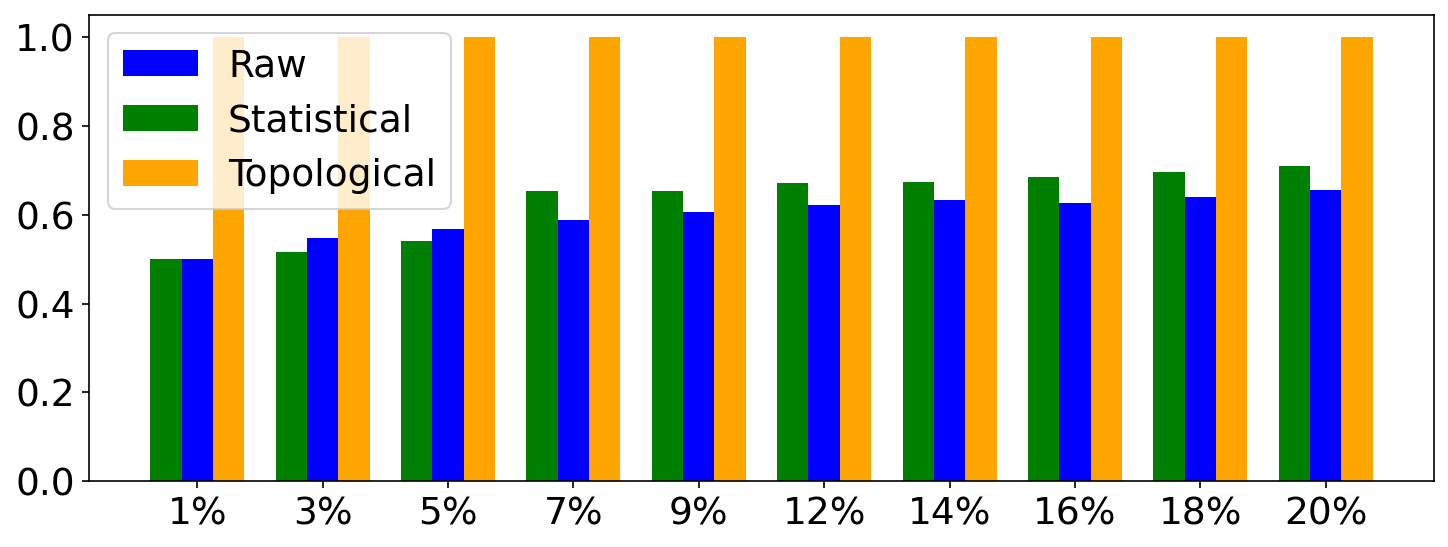}
		\caption{The area under the ROC curve from the raw, statistical and the topological features for the different unbalanced dataset with the rate from 1\% to 20\%, from the KNN algorithm .}
		\label{roc_auc_scores}
	\end{figure}

    \atabey{
    Before we present our analyses, we must note that even though the confusion matrices we present in Figure~\ref{confusionmatrix_unbalanced} may be as helpful in assessing which models and which engineered features are preferment in the unbalanced sampling scheme as in the unbalanced scheme, the areas under the ROC curves we present in Figure~\ref{roc_auc_scores} are better metrics than the accuracy scores we present in Figure~\ref{MLModels_un} for the sake of completeness. 
    }
    
    \atabey{
    Both the confusion matrices, and the area under the ROC curves for the worst performing KNN model in the unbalanced scheme collectively indicate that the models based on the proposed topological features still perform better than the models based on the statistical and raw features.
    }

\subsection{Computation Details}
    \atabey{
    Engineering topological features requires working with persistence diagrams, and constructing a persistence diagram from a point cloud is computationally expensive due to high computational cost of persistent homology. To make the calculations, we used the Turkish National Center for High Performance Computing (UHeM) at Istanbul Technical University. 
    We used the {\tt giotto-ph} library to compute the topological features in parallel. All the computations are performed on a Centos 7 Linux UHeM cluster with 128 GB RAM, Intel(R) Xeon(R) E5-2680 CPU 2.40GHz with 28 cores. 
    }
    
    \begin{table}[h]
    \centering
        \caption{}
        \label{tab:computation-cost}
        \renewcommand{\arraystretch}{1.5}
        \begin{tabular}{| l | l | l | l |}
        \hline
        \multicolumn{2}{|l|}{\textbf{Features}}                                 & \textbf{Mean} & \textbf{Std} \\ \hline
        \multicolumn{1}{|l|}{}                                       & Parallel & 45.9 s        & 321 ms       \\ \cline{2-4} 
        \multicolumn{1}{|l|}{\multirow{-2}{*}{\textbf{Topological}}} & Serial   & 111 s         & 1900 ms      \\ \hline
        \multicolumn{1}{|l|}{}                                       & Parallel  & 1.12 s        & 40.3 ms      \\ \cline{2-4} 
        \multicolumn{1}{|l|}{\multirow{-2}{*}{\textbf{Statistical}}} & Serial   & 2.10 s        & 3.16 ms      \\ \hline
        \end{tabular}
    \end{table}
    
    \atabey{
    To compute computational cost of our approaches, we sampled 50 time series  from a Cauchy process with randomly varying lengths between 500 and 1500. For each time series, we individually calculated the topological and statistical features using both serial and parallel computation. We ran 7 experiments and recorded of how long it took to compute topological and statistical features for all of the 50 time series. The results are presented in Table~\ref{tab:computation-cost}.
    }
    
    \atabey{
    The results indicate that parallel computing dramatically reduces computation time, and that the topological features are computationally more expensive than the statistical features.  In calculating the topological features, most time is spent on getting a persistence diagram from a point cloud. For this task, used {\tt giotto-ph} python library as it automatically parallelize the calculations. Despite the time complexity issues, one must recall that the topological features consistently yield more accurate results in both the balanced and unbalanced schemes. 
    }
\section{Conclusion}

	We used the raw, statistical and the topological features to classify time series sampled from different stochastic processes. In our simulation experiments we sampled times series from Wiener and Cauchy processes in both balanced and unbalanced sampling schemes. We then compared machine learning classification models built on topological features and statistical features we engineered on the sampled time series. The results show that the engineered topological features perform consistently better than statistical or raw features in building machine learning classification models even when a given dataset is unbalanced. Our experimental result show that the topologically engineered features alone can distinguish between different stochastic processes, even when statistical or raw features do not. \atabey{This means the topological feature engineering can play a critical role for time series classification tasks. Even though the result indicates that topological approach has the best accuracy, the computational cost appears as a formidable obstacle, but parallelition of calculations  appears to be useful.}
	

\subsection{Future work}

    \atabey{The main difference between the stochastic processes we used in this study, is a difference in their Levy subordinators. The experimental study we perform shows that the differences in Levy subordinators can be explained by topological features. This research question is beyond the scope of this paper, and will be an interesting direction for future work.}

    \atabey{In classification of time series obtained from Wiener and Cauchy processes},  our experiments indicate that topological features perform consistently better than standard statistical features for building machine learning models even when classes are not balanced. \atabey{One may investigate the role that topological features play in distinguishing different Levy subordinators theoretically}. One may also repeat our experiments on real world time series datasets. One must also compare our methods with other methods such as the Discrete Wavelet Transforms (DWT), the Discrete Fourier Transforms (DFT) and the Power Spectral Densities (PSD). One may also focus on topological distances between time series to apply clustering algorithms such as hierarchical clustering. 
	
\section*{Acknowledgment}

    \atabey{We thank the two anonymous reviewers whose comments helped to improve the quality of this study.}

\printbibliography

@article{FalseNearestNeighbor,
  title = {Determining embedding dimension for phase-space reconstruction using a geometrical construction},
  author = {Kennel, Matthew B. and Brown, Reggie and Abarbanel, Henry D. I.},
  journal = {Physical Review A},
  volume = {45},
  issue = {6},
  pages = {3403--3411},
  numpages = {0},
  year = {1992},
  month = {Mar},
  publisher = {American Physical Society}
}

@book{applebaum2009levy,
  title={L{\'e}vy processes and stochastic calculus},
  author={Applebaum, David},
  year={2009},
  publisher={Cambridge university press}
}

@inproceedings{chen2016xgboost,
  title={Xgboost: A scalable tree boosting system},
  author={Chen, Tianqi and Guestrin, Carlos},
  booktitle={Proceedings of the 22nd acm sigkdd international conference on knowledge discovery and data mining},
  pages={785--794},
  year={2016}
}

@book{python3,
 author = {Van Rossum, Guido and Drake, Fred L.},
 title = {Python 3 Reference Manual},
 year = {2009},
 isbn = {1441412697},
 publisher = {CreateSpace},
 address = {Scotts Valley, CA}
}

@article{geman2002pure,
  title={Pure jump L{\'e}vy processes for asset price modelling},
  author={Geman, H{\'e}lyette},
  journal={Journal of Banking \& Finance},
  volume={26},
  number={7},
  pages={1297--1316},
  year={2002},
  publisher={Elsevier}
}

@book{kijima2002stochastic,
  title={Stochastic processes with applications to finance},
  author={Kijima, Masaaki},
  year={2002},
  publisher={Chapman and Hall/CRC}
}

@book{won2004stochastic,
  title={Stochastic image processing},
  author={Won, Chee Sun and Gray, Robert M},
  year={2004},
  publisher={Springer Science \& Business Media}
}

@book{jacobs2010stochastic,
  title={Stochastic processes for physicists: understanding noisy systems},
  author={Jacobs, Kurt},
  year={2010},
  publisher={Cambridge University Press}
}

@article{tour2018cos,
  title={COS method for option pricing under a regime-switching model with time-changed L{\'e}vy processes},
  author={Tour, G and Thakoor, N and Khaliq, AQM and Tangman, DY},
  journal={Quantitative Finance},
  volume={18},
  number={4},
  pages={673--692},
  year={2018},
  publisher={Taylor \& Francis}
}

@book{schoutens2003levy,
  title={L{\'e}vy processes in finance: pricing financial derivatives},
  author={Schoutens, Wim},
  year={2003},
  publisher={Wiley Online Library}
}

@article{landriault2021analysis,
  title={On the analysis of deep drawdowns for the L{\'e}vy insurance risk model},
  author={Landriault, David and Li, Bin and Lkabous, Mohamed Amine},
  journal={Insurance: Mathematics and Economics},
  volume={100},
  pages={147--155},
  year={2021},
  publisher={Elsevier}
}

@article{kluppelberg2004ruin,
  title={Ruin probabilities and overshoots for general L{\'e}vy insurance risk processes},
  author={Kl{\"u}ppelberg, Claudia and Kyprianou, Andreas E and Maller, Ross A},
  journal={The Annals of Applied Probability},
  volume={14},
  number={4},
  pages={1766--1801},
  year={2004},
  publisher={Institute of Mathematical Statistics}
}

@misc{kats,
  author = {Facebook},
  title = {Kats},
  year = {2021},
  publisher = {GitHub},
  journal = {GitHub repository},
  howpublished = {\url{https://github.com/facebookresearch/kats}}
}

@misc{stochastic,
  author = {Flynn, Christopher},
  title = {stochastic},
  year = {2021},
  publisher = {GitHub},
  journal = {GitHub repository},
  howpublished = {\url{https://github.com/crflynn/stochastic}}
}

@incollection{nanda2014simplicial,
  title={Simplicial models and topological inference in biological systems},
  author={Nanda, Vidit and Sazdanovi{\'c}, Radmila},
  booktitle={Discrete and topological models in molecular biology},
  pages={109--141},
  year={2014},
  publisher={Springer}
}

@inproceedings{venkataraman2016persistent,
  title={Persistent homology of attractors for action recognition},
  author={Venkataraman, Vinay and Ramamurthy, Karthikeyan Natesan and Turaga, Pavan},
  booktitle={2016 IEEE international conference on image processing (ICIP)},
  pages={4150--4154},
  year={2016},
  organization={IEEE}
}

@article{chazal2021introduction,
  title={An introduction to topological data analysis: fundamental and practical aspects for data scientists},
  author={Chazal, Fr{\'e}d{\'e}ric and Michel, Bertrand},
  journal={Frontiers in Artificial Intelligence},
  volume={4},
  year={2021},
  publisher={Frontiers Media SA}
}

@article{otter2017roadmap,
  title={A roadmap for the computation of persistent homology},
  author={Otter, Nina and Porter, Mason A and Tillmann, Ulrike and Grindrod, Peter and Harrington, Heather A},
  journal={EPJ Data Science},
  volume={6},
  pages={1--38},
  year={2017},
  publisher={Springer}
}

@article{cohen2007stability,
  title={Stability of persistence diagrams},
  author={Cohen-Steiner, David and Edelsbrunner, Herbert and Harer, John},
  journal={Discrete \& computational geometry},
  volume={37},
  number={1},
  pages={103--120},
  year={2007},
  publisher={Springer}
}

@incollection{bubenik2020persistence,
	title={The persistence landscape and some of its properties},
	author={Bubenik, Peter},
	booktitle={Topological Data Analysis},
	pages={97--117},
	year={2020},
	publisher={Springer}
}

@article{adcock2016ring,
	title={The ring of algebraic functions on persistence bar codes},
	author={Adcock, Aaron and Carlsson, Erik and Carlsson, Gunnar},
	journal={Homology, Homotopy and Applications},
	volume={18},
	number={1},
	pages={381--402},
	year={2016},
	publisher={International Press of Boston}
}

@article{khasawneh2016chatter,
  title={Chatter detection in turning using persistent homology},
  author={Khasawneh, Firas A and Munch, Elizabeth},
  journal={Mechanical Systems and Signal Processing},
  volume={70},
  pages={527--541},
  year={2016},
  publisher={Elsevier}
}

@misc{truong2017exploration,
  title={An exploration of topological properties of high-frequency one-dimensional financial time series data using TDA},
  author={Truong, Patrick},
  publisher={Ph. D. thesis, KTH Royal Institute of Technology},
  year={2017}
}

@article{pereira2015persistent,
  title={Persistent homology for time series and spatial data clustering},
  author={Pereira, C{\'a}ssio MM and de Mello, Rodrigo F},
  journal={Expert Systems with Applications},
  volume={42},
  number={15-16},
  pages={6026--6038},
  year={2015},
  publisher={Elsevier}
}

@inproceedings{gidea2017topological,
  title={Topological data analysis of critical transitions in financial networks},
  author={Gidea, Marian},
  booktitle={International conference and school on network science},
  pages={47--59},
  year={2017},
  organization={Springer}
}

@article{umeda2017time,
  title={Time series classification via topological data analysis},
  author={Umeda, Yuhei},
  journal={Information and Media Technologies},
  volume={12},
  pages={228--239},
  year={2017},
  publisher={Information and Media Technologies Editorial Board}
}

@inproceedings{seversky2016time,
  title={On time-series topological data analysis: New data and opportunities},
  author={Seversky, Lee M and Davis, Shelby and Berger, Matthew},
  booktitle={Proceedings of the IEEE conference on computer vision and pattern recognition workshops},
  pages={59--67},
  year={2016}
}

@article{perea2015sliding,
  title={Sliding windows and persistence: An application of topological methods to signal analysis},
  author={Perea, Jose A and Harer, John},
  journal={Foundations of Computational Mathematics},
  volume={15},
  number={3},
  pages={799--838},
  year={2015},
  publisher={Springer}
}

@incollection{takens1981detecting,
  title={Detecting strange attractors in turbulence},
  author={Takens, Floris},
  booktitle={Dynamical systems and turbulence, Warwick 1980},
  pages={366--381},
  year={1981},
  publisher={Springer}
}

@misc{kerber2017geometry,
  title={Geometry helps to compare persistence diagrams},
  author={Kerber, Michael and Morozov, Dmitriy and Nigmetov, Arnur},
  year={2017},
  publisher={ACM New York, NY, USA}
}

@inproceedings{zhang2020gpu,
  title={GPU-Accelerated Computation of Vietoris-Rips Persistence Barcodes},
  author={Zhang, Simon and Xiao, Mengbai and Wang, Hao},
  booktitle={36th International Symposium on Computational Geometry (SoCG 2020)},
  year={2020},
  organization={Schloss Dagstuhl-Leibniz-Zentrum f{\"u}r Informatik}
}

@misc{burella2021giottoph,
      title={giotto-ph: A Python Library for High-Performance Computation of Persistent Homology of Vietoris--Rips Filtrations},
      author={Julián Burella Pérez and Sydney Hauke and Umberto Lupo and Matteo Caorsi and Alberto Dassatti},
      year={2021},
      eprint={2107.05412},
      archivePrefix={arXiv},
      primaryClass={cs.CG}
}

@article{tauzin2020giotto,
  title={giotto-tda: A topological data analysis toolkit for machine learning and data exploration},
  author={Tauzin, Guillaume and Lupo, Umberto and Tunstall, Lewis and P{\'e}rez, Julian Burella and Caorsi, Matteo and Medina-Mardones, Anibal and Dassatti, Alberto and Hess, Kathryn},
  journal={arXiv preprint arXiv:2004.02551},
  year={2020}
}

@article{sklearn,
	title={Scikit-learn: Machine Learning in {P}ython},
	author={Pedregosa, Fabian and Varoquaux, Ga{\"{e}}l and Gramfort, Alexandre  and Michel, Vincent  and Thirion, Bertrand and Grisel, Olivier and Blondel, Mathieu and Prettenhofer, Peter and Weiss, Ron and Dubourg, Vincent and Vanderplas, Jake and Passos, Alexandre and Cournapeau, David and Brucher, Matthieu and Perrot, Matthieu and Duchesnay, Édouard},
	journal={Journal of Machine Learning Research},
	volume={12},
	pages={2825--2830},
	year={2011}
}

@article{terrace2018persistence,
	title={Persistence terrace for topological inference of point cloud data},
	author={Moon, Chul and Giansiracusa, Noah and Lazar, Nicole A},
	journal={Journal of Computational and Graphical Statistics},
	volume={27},
	number={3},
	pages={576--586},
	year={2018},
	publisher={Taylor \& Francis}
}

@article{bubenik2015statistical,
	title={Statistical topological data analysis using persistence landscapes},
	author={Bubenik, Peter},
	journal={The Journal of Machine Learning Research},
	volume={16},
	number={1},
	pages={77--102},
	year={2015},
	publisher={JMLR. org}
}

@article{image2017persistence,
	title={Persistence images: A stable vector representation of persistent homology},
	author={Adams, Henry and Emerson, Tegan and Kirby, Michael and Neville, Rachel and Peterson, Chris and Shipman, Patrick and Chepushtanova, Sofya and Hanson, Eric and Motta, Francis and Ziegelmeier, Lori},
	journal={The Journal of Machine Learning Research},
	volume={18},
	number={1},
	pages={218--252},
	year={2017},
	publisher={JMLR. org}
}

@article{majumdar2020clustering,
  title={Clustering and classification of time series using topological data analysis with applications to finance},
  author={Majumdar, Sourav and Laha, Arnab Kumar},
  journal={Expert Systems with Applications},
  volume={162},
  pages={113868},
  year={2020},
  publisher={Elsevier}
}

@article{carlsson2009topology,
	title={Topology and data},
	author={Carlsson, Gunnar},
	journal={Bulletin of the American Mathematical Society},
	volume={46},
	number={2},
	pages={255--308},
	year={2009}}

@article{ghrist2008barcodes,
	Author = {Ghrist, Robert},
	Journal = {Bulletin of the American Mathematical Society},
	Number = {1},
	Pages = {61--75},
	Title = {Barcodes: the persistent topology of data},
	Volume = {45},
	Year = {2008}}

@article{chung2019persistence,
  title={Persistence curves: A canonical framework for summarizing persistence diagrams},
  author={Chung, Yu-Min and Lawson, Austin},
  journal={arXiv preprint arXiv:1904.07768},
  year={2019}
}

@article{merelli2015topological,
  title={Topological characterization of complex systems: Using persistent entropy},
  author={Merelli, Emanuela and Rucco, Matteo and Sloot, Peter and Tesei, Luca},
  journal={Entropy},
  volume={17},
  number={10},
  pages={6872--6892},
  year={2015},
  publisher={Multidisciplinary Digital Publishing Institute}
}

@article{gidea2018topological,
  title={Topological data analysis of financial time series: Landscapes of crashes},
  author={Gidea, Marian and Katz, Yuri},
  journal={Physica A: Statistical Mechanics and its Applications},
  volume={491},
  pages={820--834},
  year={2018},
  publisher={Elsevier}
}
\end{document}